# Efficient divide-and-conquer registration of UAV and ground LiDAR point clouds through canopy shape context

Jie Shao [a,b], Wei Yao [b], Peng Wan [c,*], Lei Luo [d], Jiaxin Lyu [a], Wuming Zhang [a,e,*]

[a] *School of Geospatial Engineering and Science, Sun Yat-Sen University, Zhuhai 519082, China*
[b] *Department of Land Surveying and Geo-Informatics, The Hong Kong Polytechnic University, Hung Hom, Kowloon, Hong Kong SAR, China*
[c] *Changjiang River Scientific Research Institute (CRSRI), Wuhan, China*
[d] *Key Laboratory of Digital Earth Science, Aerospace Information Research Institute, Chinese Academy of Sciences, Beijing 100094, China*
[e] *Southern Marine Science and Engineering Guangdong Laboratory (Zhuhai), Guangdong, China*

**Abstract**

Registration of unmanned aerial vehicle laser scanning (ULS) and ground light detection and ranging (LiDAR) point clouds in forests is critical to create a detailed representation of a forest structure and an accurate inversion of forest parameters. However, forest occlusion poses challenges for marker-based registration methods, and some marker-free automated registration methods have low efficiency due to the process of object (e.g., tree, crown) segmentation. Therefore, we use a divide-and-conquer strategy and propose an automated and efficient method to register ULS and ground LiDAR point clouds in forests. Registration involves coarse alignment and fine registration, where the coarse alignment of point clouds is divided into vertical and horizontal alignment. The vertical alignment is achieved by ground alignment, which is achieved by the transformation relationship between normal vectors of the ground point cloud and the horizontal plane, and the horizontal alignment is achieved by canopy projection image matching. During image matching, vegetation points are first distinguished by the ground filtering algorithm, and then, vegetation points are projected onto the horizontal plane to obtain two binary images. To match the two images, a matching strategy is used based on canopy shape context features, which are described by a two-point congruent set and canopy overlap. Finally, we implement coarse alignment of ULS and ground LiDAR datasets by combining the results of ground alignment and image matching and finish fine registration. Also, the effectiveness, accuracy, and efficiency of the proposed method are demonstrated by field measurements of forest plots. Experimental results show that the ULS and ground LiDAR data in different plots are registered, of which the horizontal alignment errors are less than 0.02 m, and the average runtime of the proposed method is less than 1 second.

## 1. Introduction

Three-dimensional (3D) structural description is an important topic in forestry inventories and is valuable when studying forest ecological processes and biodiversity. Thus, how to acquire structural information of forests completely and precisely has become an important issue that must be solved for forest ecological monitoring, inversion, and evaluation. Developments in remote sensors have enabled the precise description of forest structure. In particular, light detection and ranging (LiDAR), as an active remote sensing technology, can automatically, quickly, and precisely acquire structural information ([Luo et al., 2019](); [Shao et al., 2021a]()), which provides opportunities for precise forestry inventories.

LiDAR observes and measures forests from the ground and from above ([Guo et al., 2021]()); thus, it is generally divided into ground LiDAR and airborne laser scanning (ALS). Terrestrial laser scanning (TLS) and personal laser scanning (e.g., backpack laser scanning or BLS) are common ground LiDAR systems, of which TLS observes scenarios from the ground and can acquire details at millimeter accuracy ([Liang et al, 2016](); [Shao et al., 2020]()). More recently, ground LiDAR has been shown to be able to measure high-quality tree attributes in forests, such as tree position, diameter breast height, stem curve and leaf area index ([Chen et al., 2018](); [Pimont et al., 2018](); [Yrttimaa et al., 2020]()). However, due to occlusion from canopies, ground LiDAR has difficulty gathering complete information above the canopy and determining accurate tree height and canopy attributes in dense forests. ALS includes manned and unmanned aerial vehicle laser scanning (ULS), observes forests from the air, and acquires structural information above canopies, thus serving as a complementary technique to ground LiDAR. In general, the tree height measured by ALS is more accurate than conventional field measurements (Wang et al., 2019). However, ALS is limited in scanning the understory of dense forests due to occlusion (Brede et al., 2017). Thus, LiDAR on a single platform has difficulty meeting the requirement of measuring complete forests. The registration of ground LiDAR and ALS data is important when acquiring complete structural information and achieving accurate forestry inventories.

Point cloud registration calculates an optimal rigid transformation between two point clouds, in which a transformation consists of rotation and translation. Ground LiDAR and ALS systems commonly contain global navigation satellite system (GNSS) receivers and can produce georeferenced point clouds; thus, some methods register ground LiDAR and ALS point clouds with the help of georeferenced information (Lindberg et al., 2012; Hauglin et al., 2014). However, the occlusion of trees generally weakens or blocks the GNSS signal of the understory, particularly in dense forests (Kaartinen et al., 2015), and prevents producing accurate georeferenced data, which can lead to incorrect registration (Kelbe et al., 2017; Guan et al., 2020). In this case, the feature-based method is necessary for achieving accurate registration (Chiang et al., 2019). A common framework of the feature-based method involves coarse alignment and fine registration in two steps (Zhang et al., 2016a; Persad and Armenakis, 2017). Coarse alignment is used to build coarse correspondence and provide initial transformation parameters between ground LiDAR and ALS point clouds for the next fine registration step. Due to an inaccurate result in the coarse alignment step, a fine registration step is often used to correct aligned error and accurately register the input data.

Complex forest formations pose challenges for existing feature or descriptor-based methods, and these methods based on natural attributes generally have low execution efficiency due to complex segmentation processes. To achieve high-efficiency registration of ULS and ground LiDAR point clouds in forest scenarios, we propose a novel and efficient approach based on a divide-and-conquer strategy. According to the proposed strategy, the coarse alignment of entire point clouds is divided into ground alignment and canopy alignment, where the ground alignment is used to ensure dataset consistency in the vertical direction, and the canopy alignment is used to ensure consistency in the horizontal direction. The primary steps of the proposed method can be implemented automatically but without any artificial markers, and the divide-and-conquer strategy replaces the problem of massive 3D point cloud alignment with the problem of 2D plane alignment, which can improve alignment efficiency. Specifically, ground alignment is achieved based on the normal vector of the fitted ground plane, and canopy alignment is achieved by canopy projection image matching. For canopy image matching, we develop feature primitives based on the shape context of all canopies without tree or canopy segmentation processes, which has the potential to simplify the registration process and improve efficiency. The remainder of this paper is organized as follows. Following the introduction, Section 2 discusses related research. Section 3 describes the proposed method of point cloud registration. Section 4 presents the experimental results and a discussion of field measurements, after which conclusions are introduced.

## 2. Related research

Due to occlusions among trees, GNSS-based and marker-based methods are limited in forest point cloud registration. Thus, feature-based automatic registration is gradually becoming the focus of researcher's attention. The key to feature-based registration is to extract feature primitives in each point cloud and find the correspondence (i.e., tie points) between them. After determining tie points, some typical methods, such as singular value decomposition (SVD) (Arun et al., 1987) and unit quaternions (Horn, 1987), can be used to retrieve the transformation parameters between pairwise point clouds. In this context, the extraction and descriptor of feature primitives are important for accurate registration. Typically, feature primitives include point, line, and planar features, in which point features are often used in the feature-based registration method (Cheng et al., 2018). Feature points are typically extracted using the feature descriptor (Yang et al., 2020). However, descriptor-based methods are typically suitable for scenarios involving artificial objects with highly accurate geometric feature information, such as buildings and road signs, but might not be robust to forest scenarios with high variations in noise (Mellado et al., 2016). Alternatively, many studies extract features by exploring the spatial topologic relation between points and then calculate rigid transformation parameters based on these point features, which show high robustness to noise (Yang and Chen, 2015a; Shao et al., 2019), such as the 4-point congruent sets (4PCS) algorithm (Aiger et al., 2008). The 4PCS algorithm matches four nearly-planar points in pairwise datasets and exhibits a high efficiency and anti-noise ability but might require many runtimes. Although some improvements have been proposed (Mellado et al., 2014; Theiler et al., 2014) and generally perform well for engineered surfaces, these methods might not be appropriate for complex forest scenarios. Compared with point features, line features have stronger geometric topologies and constraints and can obtain accurate alignment results (Al-Durgham and Habib, 2014). For example, Yang et al. (2015b) extracted building outline features and fund potential correspondence based on geometric constraints and then calculated initial transformation parameters between TLS and ALS data. Cheng et al. (2018) used the 4PCS algorithm to accomplish the initial registration of TLS and ALS point clouds based on building outline features. These methods demonstrated effectiveness for regular geometric scenarios, but complex forest scenes present challenges related to reliability with these methods. Planar features contain more geometric information than point or line features and have a fast convergence speed (Shao et al., 2021b), which are commonly achieved from the ground, roof surfaces, and building facades. Therefore, planar-based registration methods are commonly suitable for indoor and outdoor scenes (Ge and Wunderlich, 2016). In complex forests, the ground generally provides reliable planar features, which means point cloud registration can obtain strong vertical constraints but might lack effective horizontal constraints.

Considering the complexity of forest formations, some studies registered point clouds acquired from different locations by exploiting the natural attributes of forests, such as tree position, canopy density, canopy height, and stem attributes (Kelbe et al., 2016; Paris et al., 2017; Brede et al., 2017; Liu et al., 2017). For example, Polewski et al. (2019) used an object-level automatic approach to register ULS and backpack LiDAR (BLS) point clouds using the quality of tree positions. They first derived tree positions by fitting cylinders or lines and segmenting a single-tree strategy and then used a distance metric to generate a feature descriptor and matched the input datasets, which performed well in real test plots. However, forest type (e.g., man-made forest) and stand density might inhibit the effectiveness of the method. Additionally, Dai et al. (2019) extracted keypoints from trees based on crown density analysis and aligned keypoints using the coherent point drift algorithm and finally registered airborne and terrestrial LiDAR datasets in boreal forests with high accuracy. These methods typically take some time to segment individual trees or crowns and reduce computational efficiency. Zhang et al. (2021) used the improved fast point feature histogram algorithm to register multiscan TLS point clouds and ULS point clouds in forest plots. Their method subsampled raw point clouds and determined correspondences by considering a large-scale geometric relation between neighboring point features. They obtained a higher efficiency than the aforementioned studies, but their coarse alignment accuracy was lower. Therefore, in complex forest environments, a high-efficiency, accurate, and automated registration method must be developed.

## 3. Methods

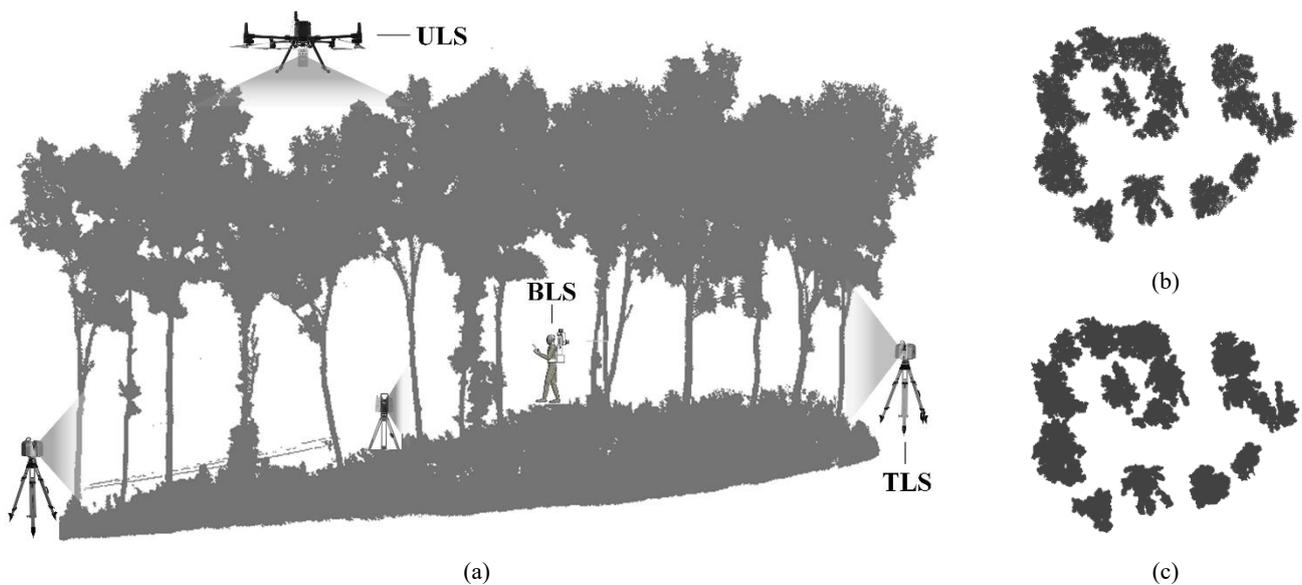

Fig. 1. Data acquisition by ULS and ground LiDAR: (a) a schematic of ULS and ground LiDAR (TLS and BLS) scanning the forest, (b) the canopy shape of the ULS point cloud in plane, and (c) the canopy shape of the TLS point cloud in plane.

Tree attributes typically exhibit certain differences due to the tree growth competition mechanism, such as different canopy shapes. Although occlusions affect the completeness of the scanned data (see Fig. 1a), LiDAR systems with penetrating capability could still acquire similar canopy shapes from the ground and air (Fig. 1b and c). Based on these premises and assumptions, we propose an automated and efficient method for registering ULS and ground LiDAR point clouds through canopy shape context features, in which ULS data are the reference and ground LiDAR point clouds are the registered data. Therefore, we present a divide-and-conquer strategy to register point clouds. Specifically, the proposed method involves three primary processes: (i) ground alignment, (ii) canopy alignment, including canopy binary image producing, preprocessing, and matching; and (iii) raw point cloud registration (Fig. 2). To ensure the consistency of two input datasets in the vertical direction, the two datasets are first aligned by adjusting the ground points to the horizontal plane (Section 3.1). Then, the two canopy binary images can be generated by projecting the vegetation points to the horizontal plane. After image preprocessing, we build feature descriptors and match canopy images of ULS and ground LiDAR data based on canopy shape context (Section 3.2). Finally, rigid transformation parameters between ULS and ground LiDAR point clouds are calculated by combining the results of ground alignment and canopy alignment (Section 3.3).

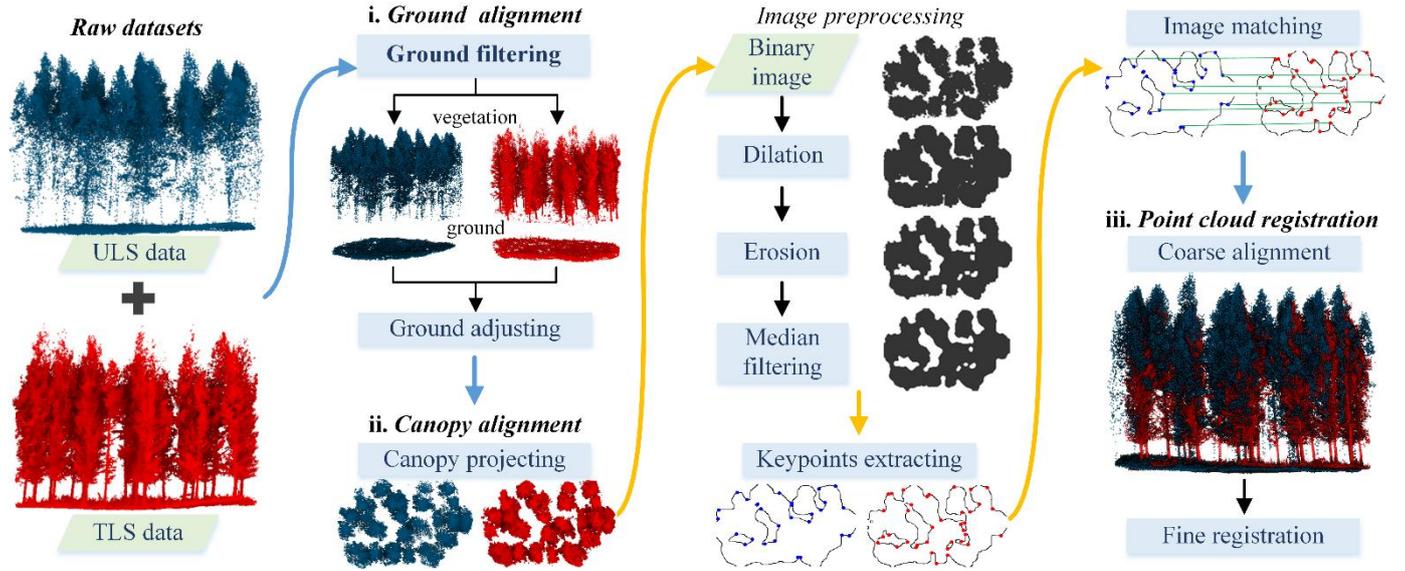

**Fig. 2.** Workflow of this paper, involving (i) ground alignment, (ii) canopy alignment (including canopy point projection, canopy binary image preprocessing, canopy contour and keypoint extraction, and image matching), and (iii) point cloud registration.

*3.1. Ground alignment*

Due to differences between the coordinate systems of ULS and ground LiDAR data, canopy shapes in binary images generated by untreated vegetation points might show inconsistencies. Therefore, to ensure the consistency of subsequent canopy binary images and improve matching accuracy, we rotate the ULS and ground LiDAR point clouds so that their corresponding grounds are parallel to the horizontal plane (i.e., xy plane of georeferenced coordinates), in which the rotation matrix is calculated based on the normal vector of the fitted ground plane. Thus, the ground point clouds of two input datasets are detected by a ground filtering process, of which the cloth simulation filtering algorithm (CSF) (Zhang et al., 2016) is used to distinguish vegetation and ground points (Fig. 3a and b). For the ground filtering algorithm, the cloth resolution is set to 0.5 m, the maximum iteration value is set to 500, and the classification threshold is set to 0.5 m.

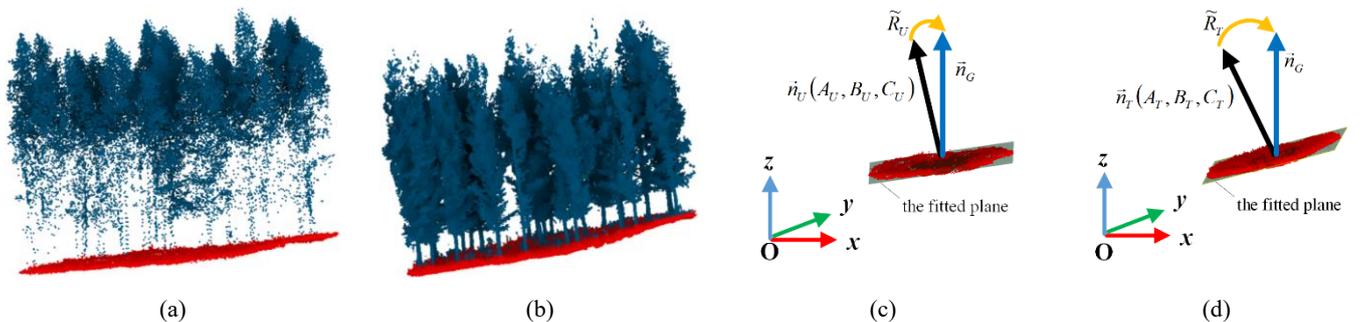

**Fig. 3.** Ground alignment. (a) and (b) show the ground filtering of ULS and TLS data, respectively (red points are grounds and blue points are vegetation); (c) and (d) show fitted ground planes of ULS and TLS data, respectively, $\vec{n}_U(A_U, B_U, C_U)$ is the ground normal vector of ULS data, $\vec{n}_T(A_T, B_T, C_T)$ is the ground normal vector of TLS data, $\vec{n}_G$ is the normal vector of the xy plane of the georeferenced coordinate, $\widetilde{R}_U$ presents the rotation matrix between $\vec{n}_U$ and $\vec{n}_G$, and $\widetilde{R}_T$ presents the rotation matrix between $\vec{n}_T$ and $\vec{n}_G$.

After determining the ground points of the two input datasets, the corresponding ground planes are fitted based on these points by the random sample consensus (RANSAC) algorithm (Fischler and Bolles, 1981). Next, we calculate an initial rotation matrix $\widetilde{R}$ based on the normal vector of each fitted plane and adjust the ULS and TLS point clouds, in which each rotation can be expressed by the Rodrigues function (Rodrigues, 1840):

$$\widetilde{R} = I + [v] + [v]^2 \times \frac{1-c}{s^2} \tag{1}$$

where $I$ is the identity matrix; $v$ is the cross product of normal vectors of the fitted ground plane and xy plane ($v_U = \vec{n}_U \times \vec{n}_G$ and $v_T = \vec{n}_T \times \vec{n}_G$); $s$ is the module of $v$; $c$ is the dot product of normal vectors of the fitted plane and xy plane; and $[v]$ is the skew-symmetric cross-product matrix of $v$ and can be defined as:

$$[v] \stackrel{\text{def}}{=} \begin{bmatrix} 0 & -v_3 & v_2 \\ v_3 & 0 & -v_1 \\ -v_2 & v_1 & 0 \end{bmatrix} (v_i \in v)$$

*3.2. Image matching based on canopy shape context*

Based on this process, ULS and ground LiDAR data can exhibit consistency in the vertical direction but not in the horizontal direction. Due to the penetrability of LiDAR, canopy shapes in ULS and ground LiDAR data commonly show similarity. Therefore, we will calculate the rotation relationship of two input datasets around the z-axis based on canopy shape context so that the ground LiDAR data are consistent with the ULS data in the horizontal direction. To improve execution efficiency, 2D image matching is used instead of 3D point cloud alignment in the horizontal direction, which involves canopy point projection, canopy binary image preprocessing, canopy contour detection, keypoint extraction and matching.

*3.2.1. Canopy binary image generation and preprocessing*

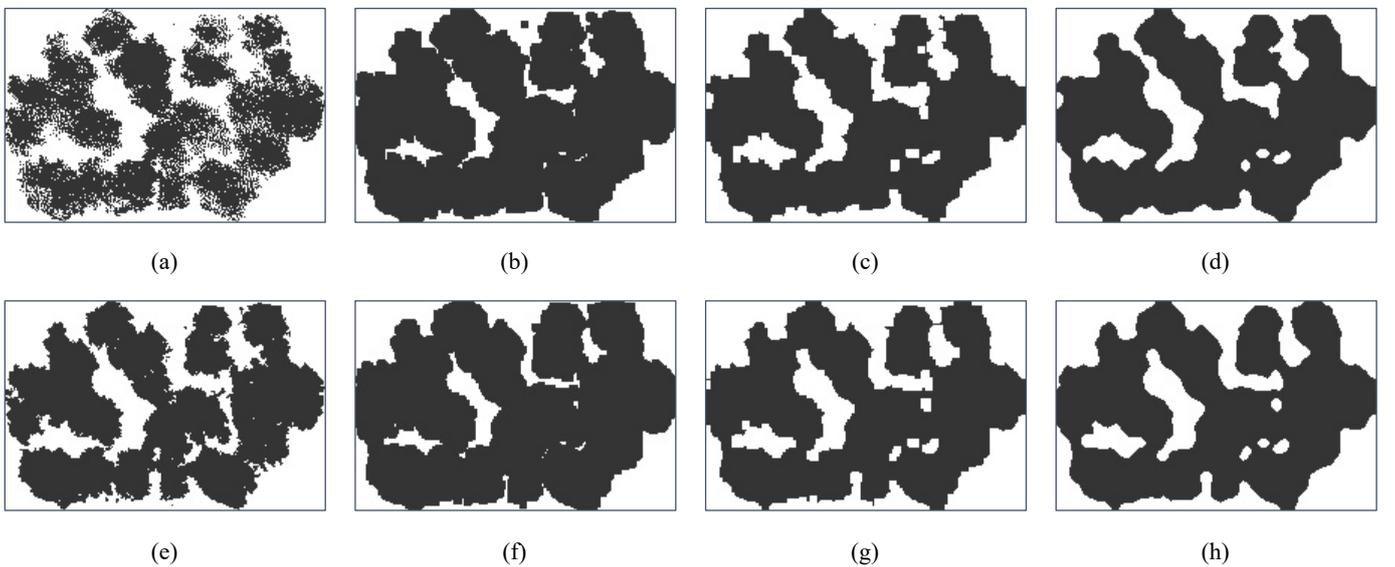

Fig. 4. Image preprocessing (black pixels represent canopy), (a)-(d) show canopy binary images of the ULS data, and (e)-(h) show canopy binary images of the ground LiDAR data; (a) and (e) present the projected canopies; (b) and (f) present dilation results; (c) and (g) present erosion results; (d) and (h) present median filtering results.

The canopy image is critical to the successful operation of the proposed method. To eliminate the influence of the understory vegetation and ground points, points 3 meters above the ground are regarded as canopy points and projected onto the xy plane at a given resolution size, in which the resolution is set to 0.1 m based on the spatial resolution of the input datasets. According to the above steps, a pair of binary images of the canopy can be obtained from the input datasets (Fig. 4); the canopy pixel value is set to 1, and the background is set to 0. Due to the discreteness and discontinuity of points, there are some holes in binary images (Fig. 4a and e); thus, dilation and erosion operations based on the morphology method are used to fill these holes (Fig. 4b, c, f, and g), in which both kernel sizes of dilation and erosion are set to five pixels. Additionally, median filtering (e.g., kernel size is set to 5 pixels) is used to smooth canopy contours and eliminate noise in canopy binary images (Fig. 4d and h). Thus, the coarse alignment of 3D point clouds is transformed to 2D image matching.

*3.2.2. Canopy contour detection and keypoint extraction*

Due to the lack of rich texture and grayscale change information in the binary images, the similarity of canopy shapes in ULS and ground LiDAR datasets is critical to image matching. To match the corresponding canopy binary images correctly, we extracted sharp corners with large geometric gradient changes on canopy contours and regarded them as keypoints for image matching. Specifically, to extract corners effectively, canopy contours are first detected by determining the connectivity $C$ between each canopy point $p_0$ and its four neighboring points $p_i$:

$$C = \frac{1}{4}\sum_{i=1}^{4} I(p_0) \cdot I(p_i) \tag{2}$$

$$p_0 \leftarrow (x, y)$$
$$p_i \leftarrow \{(x, y-1), (x, y+1), (x-1, y), (x+1, y)\}$$

where $x$ and $y$ are the pixel coordinates of the canopy contour point; $I$ represents pixel value (0 or 1); and $C$ varies between 0 and 1. If $C$ is equal to 0, $p_0$ is treated as an outlier; if $C$ is equal to 1, four neighboring points $p_{i=1,2,3,4}$ are canopy points, and $p_0$ is a canopy interior point; otherwise, there is at least one background point in the four neighboring points, and $p_0$ is a contour point (Fig. 5a).

The flat region of canopy contours generally exhibits a small curvature or slope changes, and it is difficult to provide accurate correspondences. In contrast, the sharp region commonly shows remarkable geometric characteristics. Therefore, we select the sharp corners and regard them as keypoints for canopy image matching. Because the covariance matrix can be used to explain the spread of data in the feature space, we propose extracting canopy contour keypoints using the covariance matrix based on geometric information, in which the covariance matrix $\Sigma$ of the local pixel point can be described by:

$$\Sigma = \begin{bmatrix} \sigma(x,x) & \sigma(x,y) \\ \sigma(y,x) & \sigma(y,y) \end{bmatrix} \in \mathbb{R}^{2\times 2} \tag{3}$$

where $\sigma$ is the covariance between two random variables (taking $\sigma(x, y)$ for example) and can be refined by:

$$\sigma(x, y) = \frac{1}{n-1}\sum_{i=1}^{n}(x_i - \bar{x})(y_i - \bar{y}) \tag{4}$$

where $n$ is the size of the sample points and is set based on the neighborhood radius (5 pixels in this study) of the local point, and $\bar{x}$ and $\bar{y}$ are the mean values of the sample variables. For $\Sigma$, eigenvalues of $\sigma(x,x)$ and $\sigma(y,y)$ are characteristics of the local point, where 1) if both $\sigma(x,x)$ and $\sigma(y,y)$ are large, the local point might be a corner point and will be considered a keypoint, or 2) if one of the two eigenvalues is small, the corresponding local point might be in a flat region of the canopy contour.

Additionally, inspired by the Harris corner detection approach (Harris and Stephens, 1988), we establish a response value $K$ to quantitatively identify keypoints:

$$K = det\Sigma - \alpha(trace\Sigma)^2 \tag{5}$$

where $\alpha$ is a constant (0.05 in this study), $det\Sigma$ is the determinant of $\Sigma$, and $trace\Sigma$ is the trace of $\Sigma$, in which the feature can be reflected by $det\Sigma$ and $trace\Sigma$:

$$\begin{cases} det\Sigma = \sigma(x,x)\sigma(y,y) - \sigma(x,y)\sigma(y,x) \\ trace\Sigma = \sigma(x,x) + \sigma(y,y) \end{cases} \quad (6)$$

Based on these steps, the response value $K$ of each canopy contour point is calculated. If $K$ is greater than the set threshold value, the corresponding point will be considered a keypoint (Fig. 5a).

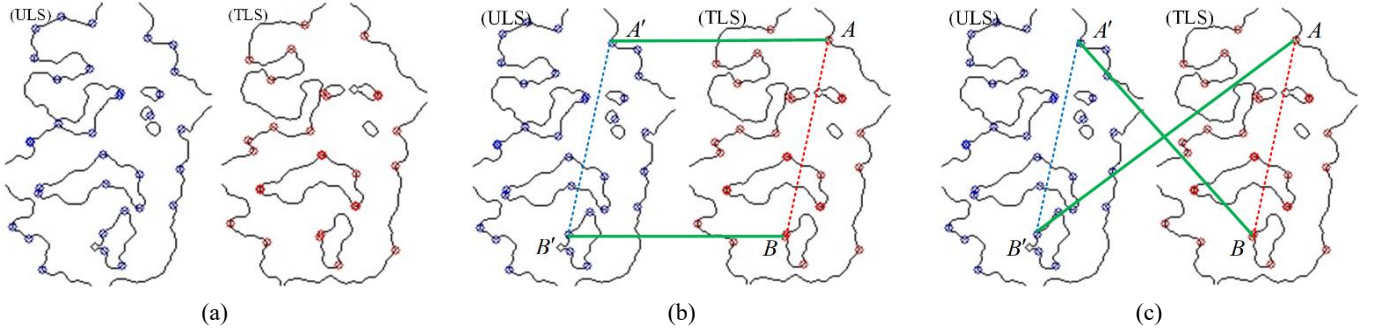

**Fig. 5.** Keypoint extraction and matching, (a) contour detection and keypoint extraction (all blue and red points), (b) correct corresponding two-point pair ($A'B'=AB$, point $A'$ and $B'$ are correspondences of point $A$ and $B$, respectively), (c) incorrect (opposite) correspondences.

*3.2.3. Image matching*

After extracting keypoints, we find corresponding keypoints and match the binary images. For two overlapping images, two accurate corresponding point pairs are generally able to achieve image matching. Due to different perspectives from the ground and air, accurate corresponding points might be difficult to find based on the keypoints' attributes. In this case, we propose a matching method based on the canopy shape context feature of the distance between two points. Therefore, a two-point congruent set containing two keypoints is created, and then, its corresponding two-point set is found within a certain distance (half of the width of the matched image). The pseudocode for this procedure is shown in Algorithm 1.

**Algorithm 1.** Finding corresponding two-point pairs

**Inputs**:
* *Keypoints* = $\{K^T\}$ *(TLS) and* $\{K^U\}$ *(ULS)*
* *Distance threshold* $\lambda$ *and* $\mu$ *(set to 5 pixels)*

**Initialize**:
* *Two-point pairs* $PT \leftarrow \emptyset$
* *Corresponding two-point pairs* $\{PP\} \leftarrow \emptyset$

**Algorithm**:
/* Producing two-point pairs in TLS */
1. **for** $i = 0$ **to** $size(K^T) - 1$ **do**
2.     **for** $j = i + 1$ **to** $size(K^T)$ **do**
       /* Distance between point $i$ and $j$ */
3.        $d_T = |K_j^T - K_i^T|$
4.        **if** $d_T > \lambda$ **then**
5.            $PT \leftarrow (K_i^T, K_j^T)$
6.        **end if**
7.     **end for**
8. **end for**
/* Searching corresponding two-point pairs */
9. **for** $i = 0$ **to** $size(K^U) - 1$ **do**
10.     **for** $j = i + 1$ **to** $size(K^U)$ **do**
11.        $d_U = |K_j^U - K_i^U|$
12.        **for** $m=0$ **to** $size(PT)$ **do**
13.            **if** $(d_U - d(PT_m)) < \mu$ **then**
14.                $PP \leftarrow \{(K_i^U, K_j^U), PT_m\}$
15.            **end if**
16.        **end for**
17.     **end for**
18. **end for**
19. **return** $PP$

According to the corresponding two-point pairs, the transformation relationship between the matched image and the reference can be established by the planar Euclidean transformation:

$$\begin{pmatrix} q_x \\ q_y \end{pmatrix} = \begin{pmatrix} \cos\theta & -\sin\theta \\ \sin\theta & \cos\theta \end{pmatrix} \begin{pmatrix} p_x \\ p_y \end{pmatrix} + \begin{pmatrix} t_x \\ t_y \end{pmatrix} \tag{7}$$

where $p$ is a keypoint in the matched image, $q$ is the corresponding keypoint of $p$ in the reference, and $\theta$ is the angle by a counterclockwise rotation. $(t_x, t_y)$ is the translation vector. The unknown rotation angle $\theta$ and translation vector can be calculated by two corresponding point pairs. However, accurate correspondences obtained by Algorithm 1 are still ambiguous, and two possible correspondences, including correct (Fig. 5b) and incorrect (or opposite) correspondences (Fig. 5c), will arise; correct correspondences between two datasets must be further determined. Thus, we analyze the overlap of two images to determine correct correspondences through the difference between pixels at the same position (Fig. 6).

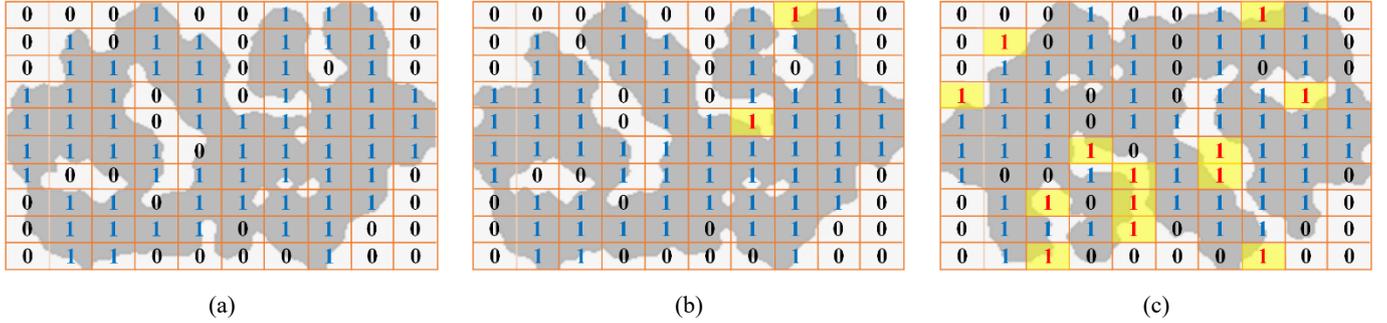

**Fig. 6.** Image overlap, (a) the matched image by gridding, 0 and 1 represent the background and canopy, respectively; (b) matching of the matched cells and the reference (canopy image of ULS data) based on correct correspondences ($O = 0.97$) (yellow cells represent incorrect correspondences); (c) image matching based on incorrect correspondences ($O = 0.8$).

To improve computational efficiency, we first grid the matched image as a block of cells (10 × 10 in this study) through a certain size (Fig. 6a), of which the value of each cell is determined by the pixel value of the cell center. Then, we compare pixel values at the same position of two images and calculate the overlap $O$ by:

$$O = 1 - \frac{1}{N} \sum_{i=1}^{N} |I(T_i) - I(U_i)| \tag{8}$$

where $I(T_i)$ is the cell center of the canopy in the matched image ($I(T_i) = 1$), $I(U_i)$ (0 or 1) is the corresponding point of $T_i$ in the reference image, and $N$ is the cell number in the matched image. If $O$ is large and trends to 1, there are correct corresponding point pairs, and a correct matching result can be obtained (Fig. 6b); if $O$ is small, there is a low overlap between the two canopy binary images, which implies an incorrect correspondence (Fig. 6c). Note that some corresponding two-point pairs can be obtained by Algorithm 1; when traversing all corresponding two-point pairs to calculate overlaps, the pair that obtains the maximum overlap will be considered the optimal correspondence for image matching.

*3.3. Registration of ULS and ground LiDAR point clouds*

Point cloud registration estimates rigid transformation parameters between ULS data $X_U$ and ground LiDAR data $X_T$, in which the transformation can be expressed by:

$$X_U = R(x, y, z) X_T + T(1:3) \tag{9}$$

where $T$ is the translation vector and $R$ is the rotation matrix ($R \in \mathbb{R}^{3\times3}$), which is calculated by rotating around the x-, y-, and z-axes of the coordinates of the ULS data. According to the ground adjustment process, the rotation parameters around the x- and y-axes have been obtained. The rotation parameter $R_{(z)}$ around the z-axis can be calculated based on the image matching result, in which $R_{(z)}$ represents ground LiDAR data rotating $\theta$ (the rotation angle of the matched image) counterclockwise around the normal vector $\vec{n}_G(0,0,1)$ of the xy plane and can be expressed by:

$$\boldsymbol{R}_{(z)} = \begin{bmatrix} \cos\theta & -\sin\theta & 0 \\ \sin\theta & \cos\theta & 0 \\ 0 & 0 & 1 \end{bmatrix} \tag{10}$$

Combining ground alignment and image matching results, rotation parameters between ground LiDAR and the adjusted ULS data can be obtained ($\boldsymbol{R}(x, y, z) = \boldsymbol{R}_{(z)}\widetilde{\boldsymbol{R}}_T$). Additionally, the translation vector $\boldsymbol{T}$ is calculated based on the midpoints of the corresponding two-point pair obtained in image matching. First, the x- and y- coordinate values of the corresponding midpoints in the spatial coordinate are estimated by their pixel coordinates and the projection resolution, and then the z-coordinate value is determined by the projection point elevation of the midpoint on the fitted ground plane. The translation $\boldsymbol{T}$ can be expressed by the difference between the spatial coordinate values corresponding to the two midpoints:

$$\boldsymbol{T} = \begin{pmatrix} x_U \\ y_U \\ -\dfrac{A_U x_U + B_U y_U + D_U}{C_U} \end{pmatrix} - \boldsymbol{R}_{(z)} \begin{pmatrix} x_T \\ y_T \\ -\dfrac{A_T x_T + B_T y_T + D_T}{C_T} \end{pmatrix} \tag{11}$$

where $(x_U, y_U)$ and $(x_T, y_T)$ are the x and y spatial coordinate values of midpoints in the adjusted ULS and ground LiDAR datasets, respectively; and $(A_U, B_U, C_U, D_U)$ and $(A_T, B_T, C_T, D_T)$ are the fitted ground planes of the two datasets. Based on these steps, we can align the raw ground LiDAR and the adjusted ULS point clouds. Note that this study is to align the raw ULS and ground LiDAR data, but not the raw ground LiDAR and adjusted ULS point clouds. Therefore, we must further rotate the aligned ground LiDAR data based on the inverse matrix $\widetilde{\boldsymbol{R}}_U^{-1}$ of $\widetilde{\boldsymbol{R}}_U$. Thus, the final transformation between the raw ground LiDAR and ULS data can be expressed by:

$$\boldsymbol{X}_U = \widetilde{\boldsymbol{R}}_U^{-1}\left(\boldsymbol{R}_{(z)}\widetilde{\boldsymbol{R}}_T \boldsymbol{X}_T + \boldsymbol{T}\right) \tag{12}$$

In general, coarse alignment has difficulty obtaining accurate registration, and a fine registration process is typically required to reduce the error in the coarse alignment step. Due to LiDAR's penetrability, there is typically an overlap between forest point clouds acquired from the ground and from the air, of which the overlap is primarily reflected in the ground and canopies. Thus, ground and canopies can provide constraints for point cloud registration in the horizontal and vertical directions, respectively. In this context, the iterative closest point (ICP) algorithm (Besl and Mckay, 1992) is selected for the fine registration of ULS and ground LiDAR point clouds.

*3.4. Evaluation criteria*

To quantify the point cloud registration performance of the proposed method, we evaluated the canopy binary image matching, coarse alignment accuracy, and fine registration accuracy results. Specifically, image matching was evaluated by the image overlap (Eq. (8)); the coarse alignment accuracy was evaluated by the minimum, maximum, average, and root mean square error (*RMSE*) of distances between corresponding features in ULS and ground LiDAR data; and the fine registration accuracy was evaluated by the minimum, maximum, average, and *RMSE* of distances between corresponding points that were selected manually, in which *RMSE* can be calculated by:

$$RMSE = \sqrt{\dfrac{\sum_{i=1}^{n} d_i^2}{n}} \tag{13}$$

where $n$ is the number of selected features and $d_i$ is the distance between corresponding features. Additionally, to evaluate the performance and advancement of the proposed method, the runtime and data performance in the coarse alignment process were analyzed and compared to other existing methods.

# 4. Results and discussion

## 4.1. Material

### 4.1.1. Study area

Two study areas were used to verify the effectiveness of this paper: one (Area 1) was located in Saihanba National Forest Park in Hebei Province in northern China, and the other (Area 2) was the Dongtai subtropical forest located in southeastern China. We selected three plots from each study area and collected data covering a range of acquisition times. Tree species included white birch (Plot 1), larch (Plot 2), Pinus sylvestris (Plot 3), poplar (Plot 4), and dawn redwood (Plot 5 and Plot 6), of which white birch and poplar are broadleaf species, and the others are coniferous species. These plots were chosen by covering a range of tree species, plot sizes, and forest structures (e.g., average diameter at breast height (DBH), average tree height, stand density, and crown density). An overview of the plot attributes is shown in Table 1.

**Table 1**
Overview of forest sample plot attributes. 'Avg.' represents the average value, and '*d*' represents the diameter of the circle.

| Area | Plot | Tree species | Sizes UAV LiDAR | Sizes Ground LiDAR | Avg. DBH (cm) | Avg. Tree Height (m) | Stand Density (n/ha) | Crown Density |
|---|---|---|---|---|---|---|---|---|
| 1 | 1 | White birch | 40 m × 32 m | 40 m × 32 m | 21.7 | 17.8 | ~150 | 0.39 |
|   | 2 | Larch | 40 m × 40 m | 25 m × 22 m | 27.1 | 19.7 | ~300 | 0.74 |
|   | 3 | Pinus sylvestris | 20 m × 15 m | 20 m × 15 m | 18.3 | 13.4 | ~900 | 0.79 |
| 2 | 4 | Poplar | Circle (*d* = 50 m) | 30 m × 30 m | 36.4 | 31.4 | 256 | 0.76 |
|   | 5 | Dawn redwood | Circle (*d* = 50 m) | 30 m × 30 m | 30.6 | 29.4 | 489 | 0.96 |
|   | 6 | Dawn redwood | Circle (*d* = 50 m) | 30 m × 30 m | 27.6 | 22.8 | 578 | 0.91 |

### 4.1.2. Data acquisition

The ULS system integrated the GNSS, inertial measurement unit (IMU), LiDAR scanner, and UAV, of which the RIEGL miniVUX-1UAV LiDAR scanner was selected and mounted on a six-rotor UAV remote sensing platform for scanning Plots 1, 2, and 3, and the Velodyne Puck VLP-16 scanner was selected to acquire data in Plots 4, 5, and 6 (Fekry et al., 2021). In addition, to verify the reliability of the proposed method, we used two ground LiDAR systems, including a TLS (RIEGL VZ-1000) system and a backpack laser scanning (BLS) system, to acquire data in Areas 1 and 2, respectively. The multiscan mode (five scans) of TLS was used to measure each plot, of which one scan was placed at the plot center, and the others were placed at four corners of the field plot. After acquiring five TLS point clouds, we registered them using a coarse-to-global strategy that was proposed in a previous study (Zhang et al., 2021). To acquire tree canopy information, a BLS system with dual scanners (Velodyne Puck VLP-16) was used to scan plots, and forest mapping was performed by integrating BLS scans and Position Orientation System (POS) information under a simultaneous localization and mapping (SLAM) framework. Table 2 lists the parameters of the ULS and ground LiDAR systems.

**Table 2**
Parameters of UAV and ground LiDAR systems, '*' indicates no reference value.

| Parameters | Area 1 ULS | Area 1 TLS | Area 2 ULS | Area 2 BLS |
|---|---|---|---|---|
| Attitude above the ground (m) | 50 | 1.5 | 86 | ~2 |
| Flight velocity (m/s) | 5 | * | 3.6 | 1 |
| Horizontal field-of-view (°) | 360 | 360 | 360 | 360 |
| Vertical field-of-view (°) | * | 100° (+60° / -40°) | 30° (+15° / -15°) | 30° (+15° / -15°) |
| Maximum range (m) | 330 | 1400 | 100 | 100 |
| Beam divergence frequency (kHz) | 100 | 1200 | 21.7 | 21.7 |
| Point density (pts/m²) | ~150 | ~20, 000 | ~85 | ~9, 600 |

## 4.2. Binary image matching results

According to the ground alignment process, ULS and ground LiDAR datasets commonly obtain consistency in the vertical direction. Then, we achieved consistency between two datasets in the horizontal direction through image matching, and to ensure matching reliability, the overlap between two canopy images was used to evaluate the effectiveness, of which the matching result corresponding to the maximum overlap was considered optimal. Therefore, we combined the matching performance and overlap to verify the image matching results (Fig. 7).

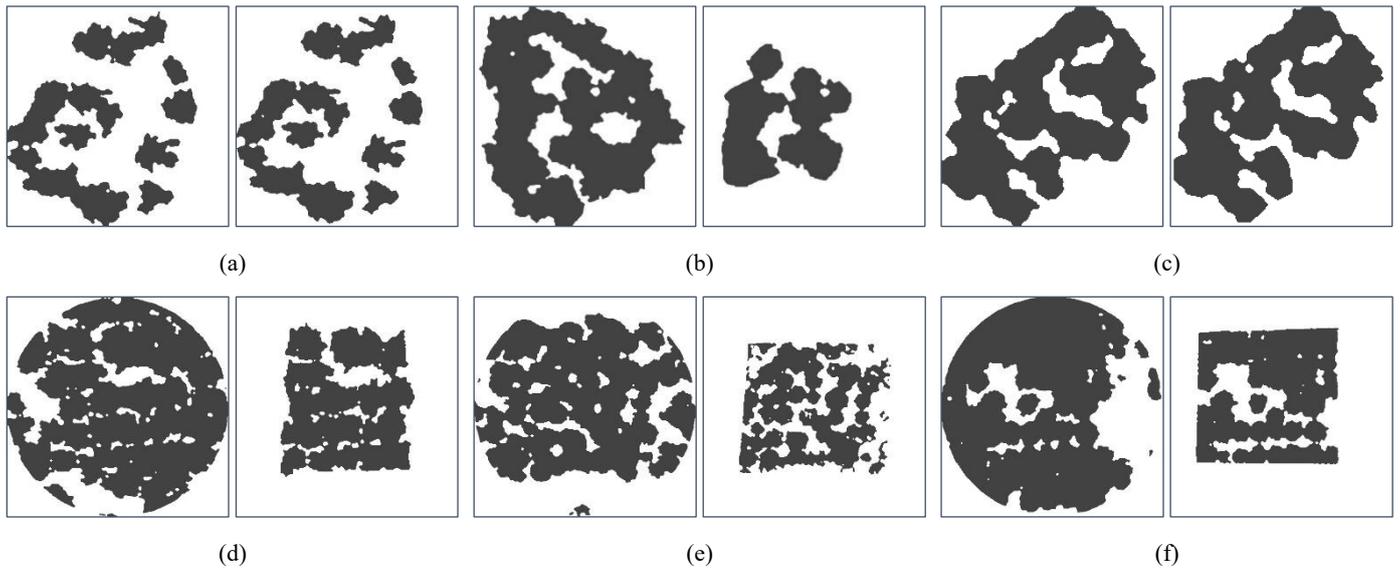

**Fig. 7.** Image matching results in field plots (left and right figures show canopy images of the ULS and ground LiDAR data, respectively, in which left canopy images are reference, right canopy images are the matched results): (a) Plot 1, the overlap is 1; (b) Plot 2, the overlap is 0.98; (c) Plot 3, the overlap is 0.97; (d) Plot 4, the overlap is 0.98; (e) Plot 5, the overlap is 0.98; (f) Plot 6, the overlap is 0.99.

As shown in Fig. 7, according to the proposed method, canopy binary images of ULS and ground LiDAR data in test plots with different attributes could be correctly matched, and overlap values were greater than 0.97, which indicated that two accurate corresponding keypoints and high overlap between canopy binary images of the ULS and ground LiDAR data were present. Thus, the image matching algorithm that combined the two-point congruent set and the canopy overlap was feasible for the test plots. In general, coarse alignment in the vertical direction could be sufficiently accurate due to the ground constraint. Therefore, the image matching result obtained by the proposed method played a decisive role in the coarse alignment and determined the effectiveness of the coarse alignment of ULS and ground LiDAR point clouds in the horizontal direction. Thus, these matching results have the potential to support reliable point cloud registration.

### 4.3. Registration results of ULS and ground LiDAR data

Combining ground adjustment and image matching results, we calculated the rotation and translation matrix between ULS and ground LiDAR point clouds and implemented coarse alignment in test plots (Fig. 8).

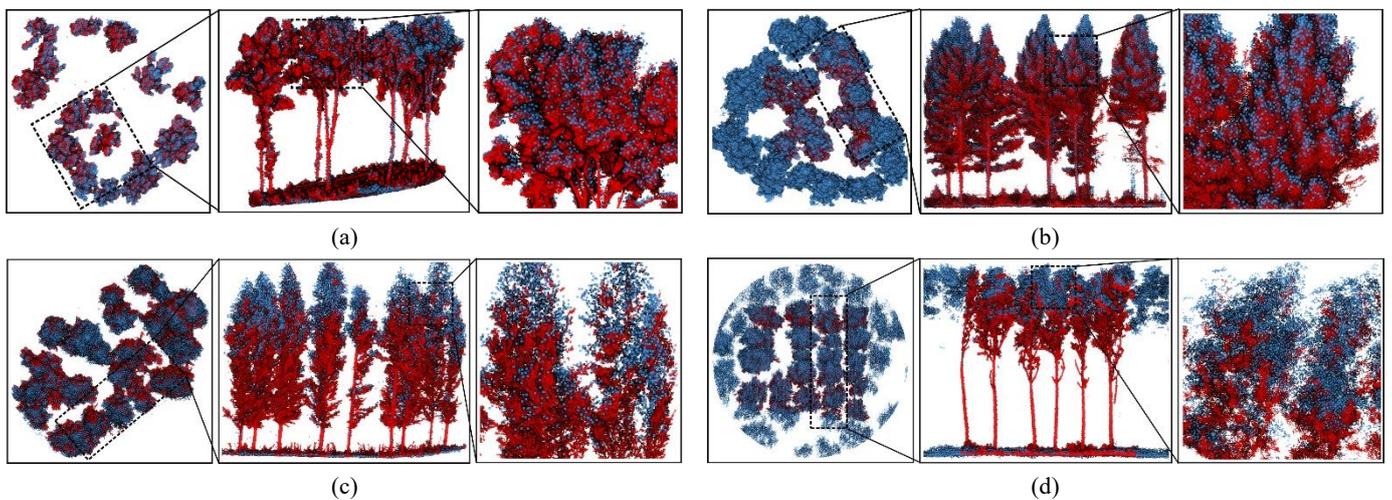

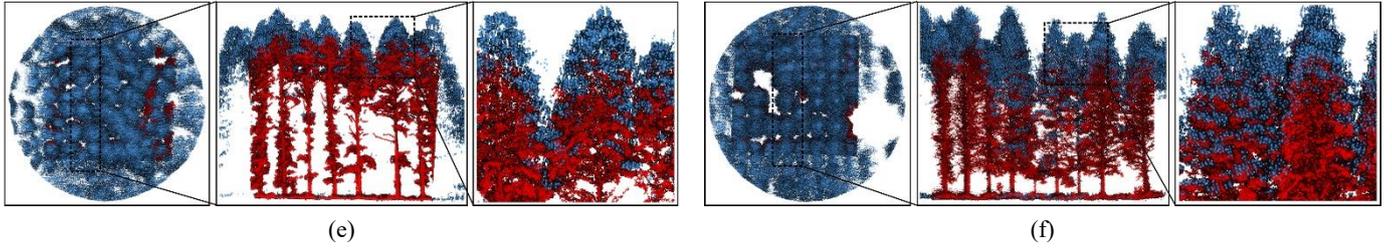

(e)                                   (f)

**Fig. 8.** Coarse alignment of ULS and ground LiDAR point clouds in test plots. ULS point clouds (blue points) are references, ground LiDAR point clouds (red points) are the aligned data, and (a)-(f) represent the coarse alignment results in Plots 1, 2, 3, 4, 5, and 6, respectively.

The coarse alignment result showed that the ground points of all ground LiDAR datasets fit well with the ground points of their corresponding ULS data by the ground alignment process, which indicated high alignment accuracy in the vertical direction. Canopy point clouds of the ground LiDAR data were also aligned with corresponding canopies of the ULS data, which explained reliable image matching and accurate point cloud alignment in the horizontal direction; the fusion of individual trees in the ULS and ground LiDAR datasets was relatively close. Overall, the attitudes of the aligned data were consistent with those of the reference data, which indicated the effectiveness of the proposed method for the coarse alignment of ULS and ground LiDAR point clouds in forest environments and had the potential to provide fine initial transformation parameters for the fine registration process of ULS and ground LiDAR point clouds (Fig. 9).

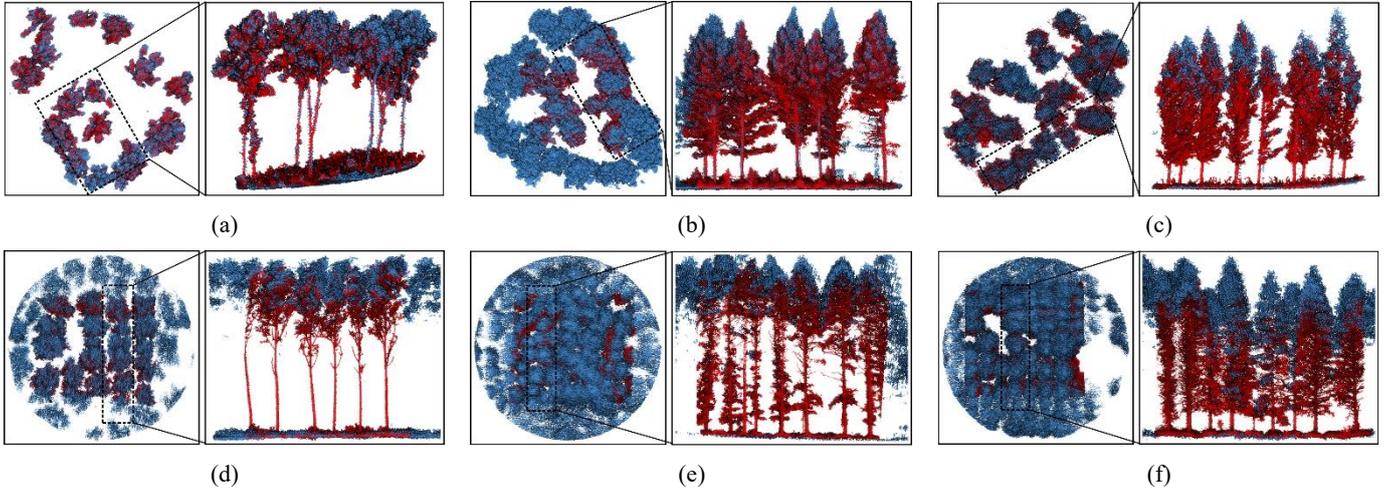

(a)                                   (b)                                   (c)

(d)                                   (e)                                   (f)

**Fig. 9.** Fine registration of ULS and ground LiDAR point clouds in test plots. ULS point clouds (blue points) are reference data, ground LiDAR point clouds (red points) are the aligned data, and (a)-(f) represent the alignment results in Plots 1, 2, 3, 4, 5, and 6, respectively.

**Table 3**
Registration accuracy. The 'N' column shows the number of the corresponding features in each plot.

| Area | Plot | N | Distance (m) in coarse alignment | | | | Distance (m) in fine registration | | | |
|---|---|---|---|---|---|---|---|---|---|---|
| | | | Min | Max | Avg. | RMSE | Min | Max | Avg. | RMSE |
| 1 | 1 | 15 | 0.06 | 0.15 | 0.11 | 0.11 | 0.04 | 0.11 | 0.07 | 0.07 |
| | 2 | 15 | 0.05 | 0.13 | 0.09 | 0.10 | 0.04 | 0.11 | 0.07 | 0.07 |
| | 3 | 15 | 0.10 | 0.26 | 0.19 | 0.20 | 0.08 | 0.13 | 0.10 | 0.10 |
| 2 | 4 | 15 | 0.09 | 0.35 | 0.20 | 0.21 | 0.05 | 0.20 | 0.13 | 0.14 |
| | 5 | 15 | 0.08 | 0.31 | 0.19 | 0.20 | 0.08 | 0.19 | 0.15 | 0.15 |
| | 6 | 15 | 0.07 | 0.24 | 0.16 | 0.17 | 0.07 | 0.16 | 0.11 | 0.11 |

Fig. 9 shows that the ICP algorithm performed well for registration of ULS and TLS point clouds in forest plots, and there was no marked registration deviation between the corresponding objects (e.g., ground and canopy). Compared with the coarse alignment result, the registered ground LiDAR data exhibited no marked changes after fine registration, which also indicated accurate coarse alignment results. Additionally, forest structural information obtained through registration of ULS and ground LiDAR point clouds was more abundant than that of a single platform dataset, which demonstrated the important of this study. To quantify the results of the coarse alignment and fine registration, we calculated the horizontal distance between corresponding features in the ULS and ground LiDAR data, with which features were detected manually. The minimum (Min), maximum (Max), average (Avg.), and RMSE of the distances are summarized in Table 3.

The distances between corresponding features in each plot varied between 0.10 and 0.35 m, and the average distances and RMSE for all plots were less than 0.21 m, which indicated effective and reliable alignment results obtained by the proposed method. After fine registration, registration accuracy improved, and the average distances and RMSE decreased to below 0.15 m. In Plots 1 and 2, the stand density and crown density were low; thus, the canopies were able to provide accurate shape context features and obtain higher coarse alignment accuracies than others, in which their alignment results were even near fine registration results. Additionally, due to high crown density and measurement error, the overall coarse alignment and fine registration accuracies in Area 2 were lower than those in Area 1.

*4.4. Robustness analysis*

Due to the complex structure of forests, automated registration of ULS and ground LiDAR point clouds is of interest in research. Tree attributes (e.g., the distance between stems) are common features for the registration of forest point clouds in existing methods. However, related methods are limited in man-made forest plots with regular layouts, in which distances between trees are similar, and it is difficult to provide accurate correspondences between two input datasets. In this study, six test plots were from man-made forests, and the layout of plots in Area 2 (Plots 4, 5, and 6) was relatively regular (Fig. 8). In contrast, due to the growth competition mechanism of trees, the canopy shapes of trees exhibited certain differences and had the potential to provide effective features for point cloud registration. From these results, the proposed method can be used for man-made forest plots with regular layouts. In addition, single tree segmentation is generally a necessary process for these methods to detect tree attributes. However, tree segmentation is easily affected by the spatial relationship between adjacent trees. Once tree canopies are connected to each other, incorrect segmentation may occur and affect point cloud registration in forests. In this study, tree canopies in each plot except for Plot 1 (Fig. 7) were almost connected, which produces challenges during tree segmentation when certain methods are used. The proposed method achieved registration using the shape context features of all canopies without the tree segmentation process to obtain the shape of a single tree canopy.

Plots with different tree species and structural parameters were selected to verify the effectiveness of the proposed method. For point clouds of the deciduous and coniferous in different areas, the proposed method can align them automatically. In the plots with different stand densities (150 n/ha, 900 n/ha), the proposed method can solve the alignment problem of ULS and ground LiDAR point clouds. Because canopy shapes provide features for point cloud registration in this paper, crown density, which is closely related to canopy shape, is a key factor affecting the effectiveness of the proposed method. In the test plots, the crown densities of Plots 1, 2, 3, and 4 were less than 0.8. Although some canopies gathered, canopy contours can still provide significant corners for accurate image matching. Thus, vegetation point clouds 3 m above the ground were projected onto the xy plane in the four plots. However, due to the large crown density in Plots 5 (0.96) and 6 (0.91), canopy shape (Fig. 9e and f), which is generated based on vegetation points 3 m above ground, makes it difficult to provide rich contour corners for image matching and leads to incorrect point cloud registration. In context, we only projected canopy points above a certain height of the ground, of which the height was set to three-quarters of the elevation difference between the highest and lowest points of the plot in this study. As shown in the experimental results, according to the above operation, plots with large crown density still showed similar canopy shapes (Fig. 7e and f) and can provide more reliable corners for achieving accurate image matching and point cloud registration. Thus, the proposed method can perform well in plots with regular layouts, different tree species, stand densities, and crown densities, which has a certain robustness.

*4.5. Data performance evaluation*

Based on the experimental results, the proposed method performed well in the test plots and achieved the coarse alignment of ULS and ground LiDAR data with an average accuracy of no more than 0.2 m. In contrast, some existing studies also solved the same problem. For example, Hauglin et al. (2014) achieved registration of ALS and TLS data using tree positions and obtained an average error of 85% and 65% of scans within 1 m and 0.5 m with ALS data with different point densities, respectively. Polewski et al. (2019) registered ULS and BLS data using tree positions and obtained a mean position deviation of 0.27-0.36 m in coniferous plots and 0.54-0.67 m in broadleaf plots. Dai et al. (2019) registered ALS and TLS point clouds through canopy density analysis and obtained an average matching distance of 0.247 m during coarse alignment. Zhang et al. (2021) aligned multiscan TLS and ULS point clouds based on the fast point feature histogram algorithm and obtained an average matching error of approximately 1 m. Compared with these existing methods, the overall coarse alignment accuracy (approximately 0.16 m in all plots) of the proposed method was higher than that of the above methods.

*4.6. Time performance analysis*

To ensure effective and accurate registration results, achieving efficient and automated point cloud registration is a key problem to be solved in this study. Therefore, to implement efficient registration of ULS and TLS data in forest scenarios, we propose a registration method based on canopy shape context features. Specifically, the proposed method was implemented on a personal computer with an Intel ® Core ™ i7-3520 M CPU @ 2.90 GHz and 8 GB of RAM, and the runtime in the coarse alignment process was recorded (see Table 5) and compared with some existing methods.

**Table 5**
Runtime performance in the coarse alignment process.

| Area | Plot | Number of LiDAR points | | Runtime (seconds) | | | |
|---|---|---|---|---|---|---|---|
| | | UAV | Ground | Ground filtering (ULS + TLS/BLS) | Ground alignment | Image matching | Total |
| 1 | 1 | 252,779 | 5,214,136 | 0.464 | 0.098 | 0.207 | 0.769 |
| | 2 | 698,577 | 3,902,277 | 0.389 | 0.097 | 0.308 | 0.794 |
| | 3 | 70,130 | 3,414,724 | 0.282 | 0.069 | 0.101 | 0.452 |
| 2 | 4 | 483,809 | 1,716,019 | 0.314 | 0.061 | 0.427 | 0.802 |
| | 5 | 452,739 | 1,557,947 | 0.354 | 0.043 | 0.428 | 0.825 |
| | 6 | 529,638 | 1,778,008 | 0.338 | 0.055 | 0.226 | 0.619 |
| | | | | | | | Average: 0.71 |

In the test plots, millions of ground LiDAR points were recorded, total runtimes in coarse alignment steps were less than 1 second, and the average runtime was 0.71 seconds. Specifically, the runtimes of the ground filtering algorithm were less than 0.5 seconds. Due to the same parameter set, the ground filtering process for the ULS data and the corresponding ground LiDAR data can be executed in parallel to reduce runtime. Except for the ground filtering process, the proposed method took less than 0.5 seconds to align ground points and achieve canopy image matching, of which the runtimes of ground alignment were less than 0.1 seconds. Image matching is critical to ensure accurate alignment of point clouds in the horizontal direction in this paper, which involves canopy image preprocessing, canopy contour detection, keypoint extraction, and corresponding two-point pair searching and matching. When the plot size and crown density are large, the image matching process generally requires more time; thus, the overall runtime of the ground alignment and image matching process in Area 2 was greater than that in Area 1. Therefore, results highlighted the high efficiency of the proposed method.

Some automated methods based on tree attributes have been presented for the registration of airborne and ground LiDAR point clouds in forest scenarios. For example, Polewski et al. (2019) presented an object-level co-registration approach based on tree position information, in which tree positions were obtained by segmenting the individual trees. Dai et al. (2019) presented an automated registration method through mode-based keypoints that were extracted from crowns, and the method generally required to segment single crowns and analyze the characteristics of crowns. Both of these methods must extract features through object (tree or crown) segmentation. However, to our knowledge, these methods are time-consuming when segmenting trees or crowns from forest point clouds; for example, the feature extraction process required 10.8 minutes on average in the study by Dai et al. (2019). Additionally, cluster trees or crowns might increase segmentation difficulty. Zhang et al. (2021) subsampled forest point clouds and directly registered ULS and TLS data without object segmentation. Although accuracy was low, efficiency was higher than that of the above two methods, and the average runtime was less than 1 minute. In contrast, the proposed method achieved coarse alignment of the input point clouds without object segmentation, obtaining high alignment accuracy and markedly improved efficiency. The average runtime of the coarse alignment was also decreased by an order of magnitude to less than 1 second.

## 5. Conclusions

To solve the problem of ULS and ground LiDAR point cloud registration and obtain complete structural information, we proposed a divide-and-conquer registration method that uses canopy shape context for forest scenarios. The method divided the coarse alignment of point clouds into two parts (ground alignment and image matching) using the dimensionality reduction strategy of converting massive point clouds into two-dimensional images and achieved the consistency of two input datasets in the vertical and horizontal directions. Additionally, for canopy binary image matching, a shape context feature based on the two-point congruent set and the canopy overlap was developed. Then, the transformation relationship between the ULS and ground LiDAR point clouds was calculated by combining the results of ground alignment and image matching. Experimental results obtained with field-measured ULS and ground LiDAR (including TLS and BLS) point clouds showed that the proposed registration method performed well in six test plots across two study areas that contained different tree species and forest structural parameters, highlighting the proposed method's reliability and robustness. The average distance between corresponding features in each test plot was less than 0.2 m, and the average runtime was less than 1 second. Comparing related existing methods, the proposed method achieved relatively high alignment accuracy and efficiency.

Canopy shape is a key factor that affects alignment accuracy in this study. However, for a plot with high crown density, canopy shapes that are described by all canopy points may not be able to adequately provide effective information for accurate alignment, requiring a reliable projection range to be assessed; in this study, this range is set based on experience. Thus, in future research, a self-adapting projection range should be developed based on crown densities at different stand heights.

## Acknowledgements

This work was supported by the National Natural Science Foundation of China (grant Nos. 41971380) and Guangxi Natural Science Fund for Innovation Research Team (grant Nos. 2019GXNSFGA245001). Our thanks also go to Lin Cao from Nanjing Forestry University, China, for kindly providing the test datasets in Area 2.